\documentclass{ecai}
\usepackage{graphicx}
\usepackage{latexsym}

\usepackage{enumitem}
\usepackage{soul}
\usepackage{url}
\usepackage[hidelinks]{hyperref}
\usepackage[utf8]{inputenc}
\usepackage[small]{caption}
\usepackage{amsmath}
\usepackage{booktabs}
\usepackage{savesym}
\savesymbol{Bbbk}
\savesymbol{openbox}

\usepackage{amssymb}
\usepackage{amsthm}
\usepackage{MnSymbol}
\usepackage{color}
\usepackage{url}
\usepackage{multirow}
\usepackage{xcolor}
\usepackage{dsfont}
\usepackage{algorithm}
\usepackage[noend]{algpseudocode}

\restoresymbol{NEW}{Bbbk}
\restoresymbol{NEW}{openbox}

\usepackage[switch]{lineno}
\usepackage{tablefootnote}

\usepackage{subcaption}
\usepackage{algorithmicx,algpseudocode}
\usepackage{array}
\usepackage{tikz} 

\newcommand{\unsim}{\mathord{\sim}}
\def\realN{\ensuremath{\mathbb{R}}}
\def\Dataset{\ensuremath{\mathcal{D}}}

\def\weightset{\ensuremath{\Theta}}
\def\Adjmat{\ensuremath{\mathbf{W}}}
\def\DAG{\ensuremath{\mathcal{G}}}
\def\DAGmap{\ensuremath{g}}

\def\feat{\ensuremath{X}}
\def\featval{\ensuremath{x}}
\def\targetval{\ensuremath{y}}

\def\featsDOM{\ensuremath{\mathcal{X}}}
\def\target{\ensuremath{Y}}
\def\targetval{\ensuremath{y}}
\def\targetDOM{\ensuremath{\mathcal{Y}}}

\def\vertSet{\ensuremath{V}}
\def\edgSet{\ensuremath{E}}

\definecolor{cadmiumgreen}{rgb}{0.0, 0.75, 0.40}
\definecolor{richcarmine}{rgb}{0.85,0.00,0.23}

\begin{document}

\begin{frontmatter}

\title{Causal Discovery and Knowledge Injection for \\ Contestable Neural Networks}

\author[A]{\fnms{Fabrizio}~\snm{Russo}%\orcid{....-....-....-....}
\thanks{Corresponding Author. Email: fabrizio@imperial.ac.uk.}
}
\author[A]{\fnms{Francesca}~\snm{Toni}%\orcid{....-....-....-....}
}
% \author[B]{\fnms{Third}~\snm{Author}\orcid{....-....-....-....}} % use of \orcid{} is optional

\address[A]{Imperial College London, UK}
% \address[B]{Short Affiliation of Second Author and Third Author}

\begin{abstract}
    Neural networks have proven to be effective at solving machine learning tasks but it is unclear whether they learn any relevant causal relationships, while their black-box nature makes it difficult for modellers to understand and debug  them.
    We propose a novel method overcoming these issues by allowing a two-way interaction  whereby 
    neural-network-empowered machines can expose the underpinning learnt causal graphs and humans can {\em contest} the machines by modifying the causal graphs before re-injecting them into the machines. The learnt models are guaranteed to conform to the graphs and adhere to expert knowledge, some of which can also be given up-front. By building a window into the model behaviour and enabling knowledge injection, our method allows practitioners to debug networks based on the causal structure discovered from the data and underpinning the predictions.
    Experiments with real and synthetic tabular data show that our method improves predictive performance up to 2.4x while producing parsimonious networks, up to 7x smaller in the input layer, compared to SOTA regularised networks.  
    
\end{abstract}

\end{frontmatter}

\vspace{-0.2cm}
\section{Introduction}

Neural Networks (NNs) have proven to be a very effective Machine Learning (ML) model for solving a wide range of problems~\cite{Goodfellow-et-al-2016}. However, it is unclear whether NNs are able to encode the Data Generating Process (DGP) and the causal structure that governs it, as shown by their weakness to data perturbation and adversarial attacks~\cite{szegedy2013intriguing,goodfellow2014explaining}. This poses a problem when deploying these models for high-stakes decisions, like granting credit or parole to individuals~\cite{rudin2019stop}.
Knowledge injection advocates the integration of human knowledge, distilled and represented in various forms, with data-driven models, including NNs (see \cite{von2021informed} for an overview).  In particular, it supports  complementing the data collected with external knowledge usually difficult to capture in the data alone. It has been shown to lead, amongst others, to less data need, improved generalisation~\cite{borghesi2020improving} and interpretability~\cite{roscher2020explainable}. In this paper, we propose a novel methodology for rendering NNs \emph{contestable} by means of  knowledge injection coupled with causal discovery.

In our methodology, knowledge injection relates to the causal structure extracted using, as a starting point, the CASTLE (CAusal STructure LEarning)~\cite{CASTLE} model, which has been shown to produce NNs better able to generalize to unseen data. For high-stakes decisions this is not enough: humans need to be in control, validate the models' recommendations and be able to challenge them. Thus, our knowledge injection methodology allows humans to inspect and modify learnt causal graphs iteratively while a disagreement remains between humans and the progressively refined model. This process can be seen as a form of contestation by the human, challenging models' outputs while still learning from data.

The need for algorithmic systems to be contestable has been advocated in AI ethics frameworks such as the ones from OECD\footnote{Principle 1.3: \url{https://oecd.ai/en/dashboards/ai-principles/P7}} and ACM\footnote{Principle 7: \url{https://www.acm.org/binaries/content/assets/public-policy/final-joint-ai-statement-update.pdf}} as well as in regulations like GDPR.\footnote{Article 22(3): \url{https://gdpr-text.com/read/article-22/}} Contestable AI has been brought to the attention of AI practitioners (see \S\ref{sec:rel_work}), however algorithmic methods for contestability are lacking.  We contribute to this landscape by enabling the interaction of Subject Matter Experts (SMEs) with NNs, aided by causal graphs which can in turn be injected into the models. The causal graphs discovered are shown to SMEs, who can cut edges deemed anti-causal or explore the most influential effects using a threshold parameter. 

Specifically,  we make the following contributions:

    \begin{itemize}
        \item We propose a first algorithm to \emph{inject}, into feed-forward NNs,  expert knowledge in the form of causal graphs. Following the injection, the NNs are guaranteed to use only the direct relationships specified in the graphs, hence adhering to the knowledge captured therein. This is a key component towards contestability, provided by our second algorithm.
        \item  We propose a second algorithm for \emph{human-AI collaboration} on the causal discovery task with NNs. This algorithm can take the human feedback on the computed causal graph underpinning the model predictions at any point of the learning process, as an iterative refinement step for model \emph{debugging} and understanding. 
        \item 
        We apply our algorithms to real and synthetic tabular data within regression and classification tasks, showing that injecting knowledge in the form of a graph can improve predictive performance up to 2.4x while making the models significantly more parsimonious (up to $85\%$ reduction in number of weights of the input layer).
        Thus, contesting a NN through its computed causal graph can increase model understanding without hurting performance.
    \end{itemize}

\vspace{-0.3cm}
\section{Related Work}
    \label{sec:rel_work}
    Von Rueden et al. \cite{von2021informed} propose a taxonomy for \emph{informed ML}, categorizing the literature on knowledge injection along three main dimensions: knowledge source, representation and integration. These capture the \emph{what} and \emph{how} of knowledge injection. Another aspect analysed is the \emph{why}.
    Our work broadly fits in the taxonomy, under \emph{Expert Knowledge} represented by \emph{Human Feedback} and integrated through the \emph{Learning Algorithm}. 
    We contribute two novel elements to the \emph{why} and the \emph{how} of knowledge injection:
    motivated by the need to support \emph{contestability}, we go beyond common incentives for expert knowledge integration into ML, such as the use of less data for training, prediction performance boosting and improved interpretability~\cite{von2021informed} and develop a novel methodology for \emph{knowledge injection empowered by causal discovery}. 
    
    In our approach, knowledge injection facilitates contestability by allowing experts to incorporate feedback into models, closing a loop that begins with showing model results to stakeholders. 
    Contestability could be seen as a prerogative of the \emph{data subject}, the receiving end of an algorithmic decision~\cite{Almada19}, who would contest the model output and, possibly, a rationale thereof. However, extending contestability to a wider range of stakeholders, including experts evaluating a decision system and professionals using it, has been recommended~\cite{henin2021beyond}.
    Existing forms of contestability range from structured interactions and model explanations~\cite{kluttz2022shaping} to normative reasoning and process modelling~\cite{TubellaTDM20}. The modality of interaction between the decision system and the ``contester'' should take into account the nature of the latter~\cite{henin2021beyond}. A data subject contesting a decision (e.g. not being granted credit) will often need layman explanations, while we focus on technical experts, providing them with detailed information in the form of causal graphs to understand and challenge model behaviour.

    Our method allows technical experts
    to contest NNs at any point of the training, entrusting them to \emph{debug} the model while validating the relationships that it is leveraging for its predictions. 
    Human-in-the-loop (HITL) training usually involves humans in data processing or annotation~\cite{wu2022survey}. Also, HITL debugging has been proposed to improve NNs used for Natural Language Processing (NLP) tasks (see \cite{lertvittayakumjorn2021explanation} for a survey). In particular, \cite{lertvittayakumjorn2020find} proposes \emph{FIND}, a method to disable ``spurious'' filters in a Convolutional NN after showing a selection of filters to practitioners in the form of word-clouds. 
    Instead of word-clouds, we employ causal graphs and we allow experts to disable spurious connections between features used in the NN. Our approach can be seen as a form of HITL contestability and debugging method guided by causal discovery. 
    
    Within the deep learning literature, in particular in the vision and NLP domains, inductive biases~\cite{goyal2022inductive} and other strategies, ranging from architecture design to weights initialisation~\cite{borghesi2020improving}, have been proposed to enhance NNs through domain knowledge. Efforts in this field have been mainly towards tweaking the loss function
    or the hyper-parameters
    to make the NNs capture known characteristic of the modelling task~\cite{borghesi2020improving}.
    Some works in this space have proposed injection of causal knowledge. Geiger et al.~\cite{geiger2022inducing} propose \emph{Interchange Intervention Training} (IIT) to induce NNs used in both computer vision and NLP tasks to have the same counterfactual behaviour of a given causal model. Zhang et al.~\cite{causal_robustness2020} propose \emph{deep CAusal Manipulation Augmented model} (CAMA), a method that uses inductive biases to make NNs robust to known manipulations of the input space, within computer vision tasks. 
    In both \cite{causal_robustness2020} and \cite{geiger2022inducing} the whole causal model is given upfront and the induction/injection works by data augmentation. Our method instead (i) focuses on the causal structure only, without making assumptions on the causal model underpinning the DGP; (ii) modifies the weights of the model and (iii) involves humans in the discovery of causal relationships from the data used for a predictive task.    
    Overall, none of the methods in the literature, to the best of our knowledge, allow for contesting discovered causal knowledge and injecting it back into NN in the form of causal graphs. 

    Beyond providing contestability and integration of expert knowledge, our method also helps in the causal discovery task (see \cite{glymour2019survey} for an overview of causal discovery methods). Meek~\cite{meek1995causal} is among the first to introduce background knowledge into the causal discovery task, however human input has been advocated since the inception of formal causal models~\cite{pearl_1988}.
    Our experiments show the benefits of injecting partial causal knowledge, without hurting and generally improving, predictive performance in the downstream task. 
    Constantinou et al.~\cite{constantinou2021impact} have investigated the impact of ten different ways to incorporate prior knowledge on causal discovery for four causal discovery methods. Chowdhury et al.~\cite{chowdhury_inducedeval} have instead analysed the impact of prior knowledge for one specific method: NOTEARS~\cite{zheng2018dags}. This is of particular interest to this paper because the method we leverage on, CASTLE~\cite{CASTLE}, includes NOTEARS' \emph{acyclicity} formulation in its loss function, to induce the discovered graph to be a DAG.
    Our experiments show that contesting models by injecting knowledge in causal form helps causal discovery in smaller data settings, thus confirming the conclusions drawn by
    both~\cite{constantinou2021impact} and~\cite{chowdhury_inducedeval}, but in a novel setting. 

 \section{Preliminaries}
    \label{sec:notation}
    
    \textit{\textbf{Modelling and Causal Set-up.}}
    Let  $\feat_1,\ldots,\feat_d$ be the set of \emph{input features} and
    $\target $ the \emph{target feature} within a regression or classification setting. Each feature $\feat_k$, for $k \in \{1, \ldots, d\}$, takes values in $\featsDOM_k\subseteq \realN$ while
    $\target$ takes values in $\targetDOM\subseteq\realN$ for regression and $\targetDOM \subset \mathbb{Z} $ for classification.
    Let $\featsDOM=\featsDOM_1 \times \ldots \times \featsDOM_d$.
    We denote with $f_\target: \featsDOM \xrightarrow[]{} \targetDOM$ a function that maps assignments of values $[\featval_1,\ldots,\featval_d]\in \featsDOM$ to 
    the input features onto a value  $\targetval \in \targetDOM$ for the target feature $\target$. In practice, $f_\target$ is learnt from a training dataset $\Dataset = \{ (\featval_{jk},\targetval_j)| j \in \{1,\ldots,N\},k \in \{1,\ldots,d\}, \featval_j \in \featsDOM,  \targetval_j \in \targetDOM \}$
    drawn from a joint distribution of values for input and target features.

    As in \cite{pearl_2009}, 
    a \emph{causal
    structure}
    over input and target features is represented by a graph $\DAG$,
    a pair
    $\langle \vertSet, \edgSet \rangle$ with
    $\vertSet = \{ \feat_1, \ldots, \feat_d, \target \}$
    the set of nodes
    and $\edgSet\subseteq \vertSet \times \vertSet$ the set of edges of $\mathcal{G}$. 
    We define a \emph{full} DAG as a causal graph that is directed, acyclic and captures all applicable causal relationships amongst \emph{all} features. For full DAGs, if an edge between two features is present (resp. absent), then the features are (resp. are not) in a causal relationship. 
    
    Given the scarcity of full DAGs for real-world applications, we also use what we call \emph{partial} causal graphs, of the form $\DAG_p = \langle \vertSet_p, \edgSet_p \rangle$ with
    nodes $\vertSet_p \subseteq \vertSet$  and edges $\edgSet_p \subseteq \vertSet_p \times \vertSet_p$. Intuitively, if $i,k \in \vertSet_p $ but $(i,k) \not\in \edgSet_p$ and $(k,i) \not\in \edgSet_p$, then node $i$ is definitely not causally related to node $k$; if $(i,k) \in \edgSet_p$ and $(k,i) \notin \edgSet_p$,
    then $i$ \emph{can} be a cause of $k$, but not vice versa. 
    Hence, if a directed edge is present but the one between the same nodes with opposite direction is absent, then the latter relation is deemed as anti-causal. 
    If $i,k \in \vertSet \setminus \vertSet_p$, then both $(i,k)$ and $(k,i)$ $\in \edgSet_p$, indicating the lack of knowledge about any causal relation among nodes $i$ and $k$.   
    Thus, if the graph is \emph{complete}~\cite{pearl_2009}, i.e. with exactly $\vertSet_p =\vertSet$  and $\edgSet_p = \vertSet \times \vertSet$, then we know that 
    all features could be causally related but not in which direction: NNs are used to aid the decision on the direction. Overall, our partial graphs compactly represent sets of constraints, and differ from full DAGs giving instead the causal structure among all observed features. 

    \vspace{0.1cm}
    \textit{\textbf{CASTLE \cite{CASTLE}}}.
    \label{sec:castle} 
    Our proposed algorithms build upon the architecture of~\cite{CASTLE}, whose schematic
    is provided in Fig.~\ref{fig:nn_structure_question}. CASTLE operates with a feed-forward NN combining $d+1$ sub-networks, each with $d$ input neurons, amounting to all input and target features minus one: each sub-network masks a distinct element amongst $\feat_1,\ldots,\feat_d,\target$. The output layer of a sub-network with feature $F$ masked in the input layer, has $F$ as output neuron. Thus, each sub-network is responsible for reconstructing one feature, without using that feature. All sub-networks have $M+1$ layers, and differ only in the input and output layers, i.e. layers $2, \ldots, M$ are shared. Hence, during training, the hidden layers are optimised to achieve the best performance for all sub-networks at the same time, thus encoding the structure of the interactions among all features~\cite{CASTLE}.

\definecolor{green1}{RGB}{19, 138, 20}
\definecolor{green2}{RGB}{63, 179, 34}
\definecolor{green3}{RGB}{78, 250, 65}
\definecolor{blue}{RGB}{0 110 175}

\def\layersep{3cm}
\def\ninput{4}
\def\nhidden{4}

\begin{figure}[t]
    \centering
    \scalebox{0.8}{
        \begin{tikzpicture}[shorten >=1pt,->,draw=black!50, node distance=\layersep]
        \scriptsize
            \tikzstyle{every pin edge}=[<-,shorten <=1pt]
            \tikzstyle{neuron}=[circle,fill=black!25,minimum size=12pt,inner sep=0pt]
            \tikzstyle{neuron1}=[circle,fill=black!25,minimum size=16pt,inner sep=0pt]
            \tikzstyle{neuron0}=[circle,fill=white!10,minimum size=12pt,inner sep=0pt]
            \tikzstyle{input neuron mask}=[neuron, fill=black!50];
            \tikzstyle{input neuron 1}=[neuron, fill=green1];
            \tikzstyle{input neuron 2}=[neuron, fill=green2];
            \tikzstyle{input neuron 3}=[neuron, fill=green3];
            \tikzstyle{output neuron}=[neuron1, fill=red!50];
            \tikzstyle{hidden neuron}=[neuron1, fill=blue!50];
            \tikzstyle{annot} = [text width=7em, text centered, minimum size=9pt]
        
            % Draw the input layer nodes 1
            \node[input neuron mask, pin=left:{\normalsize Y}] (I-1) at (0,-1/2) {};
            \foreach \name / \y in {$X_1$/2, $X_d$/4}
            % This is the same as writing \foreach \name / \y in {1/1,2/2,3/3,4/4}
                \node[input neuron 1, pin=left: \normalsize \name] (I-\y) at (0,-\y/2) {};
            \node[neuron0] (I-3) at (0,-3/2) {};
            \node (dots1) at (0,-2.8/2)  {\large $\vdots$};
  
            % Draw the input layer nodes 2
            \node[input neuron mask, pin=left:\textcolor{black!60}{\normalsize $X_1$}] (I-7) at (0,-7/2) {};
            \foreach \name / \y in {Y/6,$X_d$/9}
            % This is the same as writing \foreach \name / \y in {1/1,2/2,3/3,4/4}
                \node[input neuron 2, pin=left:\textcolor{black!60}{\normalsize  \name}] (I-\y) at (0,-\y/2) {};
            \node[neuron0] (I-8) at (0,-8/2) {};
            \node (dots2) at (0,-7.8/2)  {\large $\vdots$};
            
            % Draw the input layer nodes 3
            \foreach \name / \y in {Y/11,$X_1$/12}
            % This is the same as writing \foreach \name / \y in {1/1,2/2,3/3,4/4}
                \node[input neuron 3, pin=left: \textcolor{black!60}{\normalsize  \name}] (I-\y) at (0,-\y/2) {};
            \node[input neuron mask, pin=left:\textcolor{black!60}{\normalsize  $X_d$}] (I-14) at (0,-14/2) {};
            \node[neuron0] (I-13) at (0,-13/2) {};
            \node (dots3) at (0,-12.8/2)  {\large $\vdots$};

            % Draw the hidden layer nodes
            \foreach \name / \y in {1,...,\nhidden}
                \path[yshift=-0cm]
                    node[hidden neuron] (H-\name) at (\layersep,-\y*1.5 cm) {};
                    % node[output neuron, pin={[pin edge={->}]right:\textcolor{black!60}{\name}}] (O-\y) at (\layersep*2,-\y*1.5) {};
                    
            \node (dots) at (\layersep,-3.65)  {\large $\vdots$};
            \node (dots) at (\layersep,-5.15)  {\large $\vdots$}; 
            
            % Draw the output layer node
            % \node[output neuron,pin={[pin edge={->}]right:Output}, right of=H-3] (O) {};
            \node[output neuron, pin={[pin edge={->}] right:\normalsize $Y$}] (O-1) at (\layersep*2,-1*1.5) {};
            \foreach \name / \y in {$X_1$/2,$X_k$/3,$X_d$/4}
            % This is the same as writing \foreach \name / \y in {1/1,2/2,3/3,4/4}
                \path[yshift=-0cm]
                    node[output neuron, pin={[pin edge={->}]right:\textcolor{black!60}{\normalsize \name}}] (O-\y) at (\layersep*2,-\y*1.5) {};
            \node (dots) at (\layersep*2,-3.65)  {\large $\vdots$};
            \node (dots) at (\layersep*2,-5.15)  {\large $\vdots$}; 
            
            % Connect every node in the input layer with every node in the
            % hidden layer.
            \foreach \source in {2,3,4}
                \foreach \dest in {1,...,\nhidden}
                    \path (I-\source) edge[thick,black!80] (H-\dest);
        
            \foreach \source in {6,8,9,11,12,13}
                \foreach \dest in {1,...,\nhidden}
                    \path (I-\source) edge (H-\dest);
        
            % Connect every node in the hidden layer with the output layer
            \foreach \source in {1,...,\nhidden}
                \foreach \dest in {1}
                \path (H-\source) edge[thick,black!80] (O-\dest);
        
            \foreach \source in {1,...,\nhidden}
                \foreach \dest in {2,...,\ninput}
                \path (H-\source) edge (O-\dest);
        
            % Annotate the layers
            \node[above of=I-1, node distance=0.58cm] (il)  {\large Input layer};
            \node[right of=il] (hl) {\large Hidden layer(s)};
            \node[right of=hl] {\large Output layer};
            
            % Question annotation
            \node[align=left, color=green1, fill=white] (question) at (1.7,-0.95) {\large ($X_1, Y) \in \DAG?$};
            \foreach \dest in {1,...,\nhidden}
                \path (I-2) edge[thick, green1] (H-\dest);
                
            % Annotate the Sub-Networks
            \node[annot] (sub1) at (-1.8,-1.2)  {\large \centering Sub-Network 1};
            % \node (dots) at (-1.8,-2.4)  {\large $\vdots$};
            \node[annot,black!60] (sub2) at (-1.8,-3.7)  {\large \centering  Sub-Network 2};
            \node (dots) at (-1.8,-4.9)  {\large $\vdots$};         
            \node[annot,black!60] (sub3) at (-1.8,-6.2)  {\large \centering  Sub-Network $d+1$};

        \end{tikzpicture}
        }
    \caption{Joint Neural Network Structure. Darker arrows refer to the highlighted Sub-Network 1, predicting $Y$ (Output layer), while having $Y$ masked (grayed out in Input layer). Green arrows represent the weights that we consider masking when injecting causal knowledge answering questions like: ``is $X_1$ a parent of $Y$?''.
    }
    \label{fig:nn_structure_question}

\end{figure}
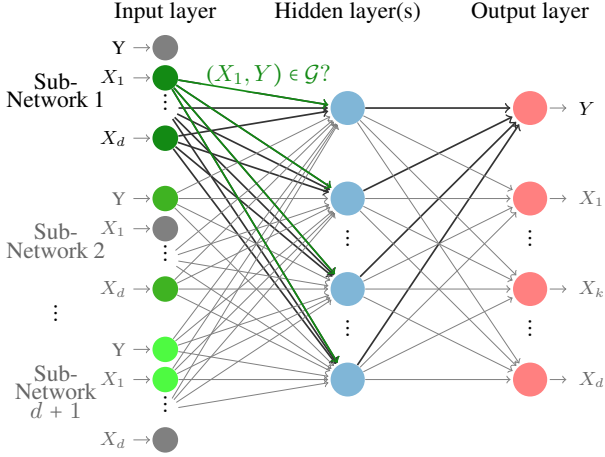

    We refer to the feed-forward NN, with all its sub-networks, as the \emph{joint NN}. We refer to weights for layer $l\in \{1, \ldots, M\}$ as
    $\Theta_l$, where $\Theta_l^{i,j}$ is the weight from neuron $i$ in layer $l$ to neuron $j$ in layer $l+1$. $\Theta_1^{i,j,k}$ stands for the weight from input neuron $i$ in layer $1$ to neuron $j$ in the first hidden layer of the $k$-th sub-network. 

    The joint NN carries out both the prediction of the target feature and the reconstruction of the input features. 
    To train the NN, like~\cite{CASTLE}, we use back-propagation and stochastic gradient descent applied to a loss function that includes a causal discovery element borrowed from NOTEARS~\cite{zheng2018dags}. 
    More precisely, the loss is formed by two modules: the prediction loss and the \emph{DAG loss}. The former is  Mean Squared Error (MSE) or cross-entropy loss for regression and classification, respectively. The DAG loss is from~\cite{zheng2018dags} and can itself be broken down into three components: the reconstruction loss, MSE for each sub-network's output, assuming they are continuous; the \emph{acyclicity} loss, a term that is 0 when \Adjmat{} (described next) represents a DAG (from Theorem 1 of \cite{zheng2018dags}); and finally an $L_1$ loss to induce sparsity in the weights' matrix.

    \Adjmat{} is a square hollow matrix of order $d+1$ holding a non-negative weight for each feature, including the target, from and towards all the others. The $(i,k)$-th entry of \Adjmat{}, for any $i,k \in \{1, \ldots, d+1\}$, results from the square root of the sum of squared input layer weights across the hidden neurons. We denote the entries of \Adjmat{} as $w_{ik}$. Formally, for $h$ the number of hidden neurons in the first hidden layer:

    \begin{equation}
        w_{ik} = \sqrt{\sum_{j=1}^{h}\left(\Theta_1^{i,j,k}\right)^2} 
        \label{eq:adjmat}
    \end{equation}
    Given the standardized input data, $w_{ik}$ represent the magnitude of the effect that each feature $i$ has on $k$. However, as discussed previously, \Adjmat{} assumes causal connotations due to the \emph{acyclicity} part of the loss function, which induces \Adjmat{} to represent a DAG. 

\section{Methodology}
    \label{sec:method}

    In this section we first introduce our algorithm to inject causal knowledge in the form of a graph into feed-forward NNs (Alg.~\ref{alg:inject_alg}). The ability to make the NN respect external assumptions about the structure of the data, as afforded by  Alg.~\ref{alg:inject_alg}, represents a key step in making NNs contestable. The contesting process is then provided in Alg.~\ref{alg:refine_alg}, which enables practitioners to challenge the recommendations of the NN, based on the causal graph discovered while computing them. Alg.~\ref{alg:refine_alg} uses Alg.~\ref{alg:inject_alg}, to inject practitioners' feedback into NNs.     

    \textit{\textbf{The Graph Injection Algorithm.}}
    \label{sec:algo}
    Alg.~\ref{alg:inject_alg}
    takes three main inputs: a training dataset \Dataset{}, a joint NN with weights $\weightset_t$, which can be randomly initialised or already fitted on \Dataset{}, and a causal graph \DAG{}. 
    The input graph \DAG\ can take two forms: a full  DAG, or a partial graph. In practice, having a full DAG is rare, hence we allow for \DAG\ to be a partial graph (see \S\ref{sec:notation}), representing hard causal constraints while allowing the discovery of additional causal relations.
    
    The output of Alg.~\ref{alg:inject_alg} amounts to a masked
    joint NN with weights $\weightset_\DAG$, which only uses the relationships contemplated in 
    \DAG{}: we call this an \emph{Injected} NN. 
    The injected NN resulting from  Alg.~\ref{alg:inject_alg} is  fitted, or tuned if the input NN has already been fitted, to use only the relationships that were deemed causal by including them in \DAG.
    The injection is achieved through the masking of the weights of the input layer without a counterpart in the input DAG, namely  $(i,k) \notin E$ means that $X_i$ cannot cause $X_k$, hence $\Theta_{1}^{i,j,k}=0$ $\forall j \in \{1,\ldots,h\}$ (see line 4 of Alg.~\ref{alg:inject_alg}) which results in $w_{ik}=0$.
    The $\texttt{update}$ function represents a back-propagation pass and will iteratively refine the NN's weights to be effective in the prediction and reconstruction tasks without using spurious anti-causal relationships.
    The final weights $\weightset_\DAG$ of the injected NN result from progressive changes, for a number of steps that is at most $T$, if a \emph{patience} threshold $T_s < T$ of loss improvement on the validation set is not reached. 
     
    \begin{algorithm}[b]
        \caption{Inject Causal Graph}
        \label{alg:inject_alg}
        \begin{flushleft}                           
            \textbf{Input}: Training Data $\mathcal{D}$; NN with $M$ layers, $h$ neurons in the first hidden layer, trained for $t$ steps with final weights $\Theta_t$; max number of steps $T$; patience $T_s<T$; causal graph $\DAG \!=\! \langle V, E \rangle$ \\
            \textbf{Function}: \texttt{inject\textunderscore graph}($\Dataset, \Theta_t, T$, $T_s, \DAG$):\\
                \vspace{-0.3cm}
        \end{flushleft}                           
        \begin{algorithmic}[1]
            \For{$i \in \{1, \ldots, d+1\}, \, \textbf{for} \, k \in \{1, \ldots, d+1\}$}, 
                \For{$j \in \{1,\ldots, h\}$}            
                    \If{$i,k \in V \And$ $(i,k) \notin E$ }
                        \State $\Theta_{1,t+1}^{i,j,k} \leftarrow 0$ \Comment{mask anti-causal relations} 
                    \Else%[]
                        \State $\mathcal{L}_{\text{best}}, t_s \leftarrow \infty, 0$
                        \While{$t < T \And t_s < T_s$}
                                \State $\Theta_{1,t+1}^{i,j,k}  \!\!\leftarrow \!\! \texttt{update}(\Theta_{1,t}^{i,j,k}, \Dataset)$\Comment{causal relations}
                                \State $\Theta_{m,t+1} \!\!\leftarrow \!\! \texttt{update}(\Theta_{m,t}, \Dataset)$ $\forall m \! \in \! \{1,\ldots, M\}$\!\!
                                \State 
                                $t \leftarrow t+1$
                                \If{$\mathcal{L}_{t}
                                < \mathcal{L}_{\text{best}}$}
                                    \State $\mathcal{L}_{\text{best}}, t_s \leftarrow \mathcal{L}_t, 0$
                                \Else
                                    \State $t_s \leftarrow t_s+1$
                                \EndIf
                                
                        \EndWhile   
                    \EndIf
                \EndFor
            \EndFor                

            \State $\Theta_\DAG \leftarrow \Theta_t$\\
            \Return 
            $\Theta_\DAG$
        \end{algorithmic}
            \vspace{-0.3cm}
        \begin{flushleft}                           
            \textbf{Output}: Injected NN with weights $\Theta_\DAG$
        \end{flushleft}                           
    \end{algorithm}
    
    Computationally, the main difference of Alg.~\ref{alg:inject_alg} from CASTLE's training loop is that our algorithm focuses the training on the weights for the accepted edges, i.e. the causal relationships. Another way of understanding this process is as a ``selective'' causal dropout method. Intuitively, the original CASTLE methodology regularises the underlying NN so that the fitted model uses causal parents \emph{more than} children and siblings for its predictions. 
    Instead of only \emph{preferring} the use of parents, our Alg.~\ref{alg:inject_alg} \emph{enforces} it: we reconstruct each feature and carry out the target prediction using \emph{only} each feature's parents.
    With this restriction, we aim at avoiding the use of a feature's children and/or siblings that may have unstable relationships with the parent feature being predicted: a change in a children/siblings will not necessarily change the feature, while a change in a parent of the feature will. 
    Effectively, we encode the answers to causal questions into our masking scheme. An example of such questions is in Fig.~\ref{fig:nn_structure_question}: \emph{does the edge from $X_1$ to $Y$ belong to our agreed causal structure $\DAG$?} If not, we set the corresponding weights to 0 and prevent the effect of $X_1$ on $Y$. Note that we mask inputs and therefore direct effects while leave indirect effects to be captured by the hidden layers.
    
    \begin{algorithm}[t]
        \caption{Contest Computed Causal Graph}
        \label{alg:refine_alg}
      \begin{flushleft}                           
            \textbf{Input}: Training Data \Dataset; NN with $M$ layers, $h$ neurons in the first hidden layer, trained for $t$ steps with final weights $\Theta_t$; 
            number of total training steps $T$; patience $T_s<T$;  
            \textit{Expert Knowledge}
            \\        
            \textbf{Function}:~\texttt{contest\textunderscore graph}($\Dataset, \Theta_t, T, T_s, \textit{Expert Knowledge}
            $):\\
                \vspace{-0.3cm}
        \end{flushleft} 
        \begin{algorithmic}[1]

            \State \textit{contested, $\tau \leftarrow$ True, $0$}
            \While{\textit{contested = True}}
                \State $\DAG \leftarrow g_{\tau}(\Adjmat_{\Theta_t})$
                \State $\DAG_r, \tau \leftarrow \texttt{revise\textunderscore graph}(\DAG, \textit{Expert Knowledge})$                  
                \If{$\DAG_r\neq \DAG$}
                    \State $\Theta_t \leftarrow $\texttt{inject\textunderscore graph}($\Dataset, \Theta_t, T, T_s, \DAG_r$)
                    \Else
                        \State \textit{contested $\leftarrow$ False}
                \EndIf
            \EndWhile
                \State $\Theta_\DAG \leftarrow \Theta_t$\\
            \Return $\DAG$, $\Theta_\DAG$
       \end{algorithmic}
            \vspace{-0.3cm}
        \begin{flushleft}                           
          \textbf{Output}: Refined DAG $\DAG$ and Injected NN with weights $\Theta_\DAG$
        \end{flushleft}
    \end{algorithm}

    \textit{\textbf{The Contesting Algorithm}.}
    \label{sec:partial_know}
    Alg.~\ref{alg:inject_alg} enables the injection of a graph representing the structural relationships among all or a subset of the features used by a NN to predict a target.
    With Alg.~\ref{alg:refine_alg} we expose, in the form of a graph, the relationships that the NN has found in the data so that practitioners can critique and contest the output. We then use Alg.~\ref{alg:inject_alg} to close the ``contestation loop'' and incorporate the human feedback into the NN.
    Alg.~\ref{alg:refine_alg} has the same inputs as Alg.~\ref{alg:inject_alg}, apart from the input graph \DAG\ which is replaced by what we call \emph{Expert Knowledge}. This represents knowledge external to the data, and provided by SMEs that have prior experience with the modelling task. To leverage this external knowledge, we need to engage with the SMEs and we do so by means of the causal graph underpinning CASTLE's predictions. 
    After calculating the adjacency matrix $\Adjmat_\weightset$ from the joint NN's weights \weightset\, using Eq.~\ref{eq:adjmat}, we transform it into a DAG $\DAG=\DAGmap_\tau(\Adjmat_\weightset)$ using the following equations:
    \begin{equation}
     \label{eq:direction_tau_edges}
        \edgSet_\tau(\Adjmat_\weightset) = \{(i,k) | w_{ik} > w_{ki} \wedge w_{ik} > \tau \} 
    \end{equation}
    \vspace{-0.4cm}
    \begin{equation}
    \label{eq:direction_tau}
        \DAGmap_\tau(\Adjmat_\weightset)= (\vertSet,\edgSet_\tau(\Adjmat_\weightset))
    \end{equation}
    Eq.~\ref{eq:direction_tau_edges}  is a simple ``edge creation function'', applied to the adjacency matrix to produce the edges of a DAG with nodes $\vertSet = \{ \feat_1, \ldots, \feat_d, \target \}$ as per Eq.~\ref{eq:direction_tau}.
    In general, the threshold $\tau\geq 0$ is meant 
    to cut out the uninfluential relationships in the data. Thus, setting $\tau=0$ results in treating all identified relationships, as represented by the elements $w_{ik}$ of $\Adjmat_\weightset$, as influential. 
 
    Having extracted a DAG $\DAG$ from the joint NN, using Eq.~\ref{eq:adjmat} to \ref{eq:direction_tau} with $\tau=0$, we present it to the SMEs for them to assess it, through the \texttt{revise\textunderscore DAG} function. The output of this revision by the SMEs can be in regard to the threshold $\tau$, to cut out more or less of the least influential effects, and/or in regard to specific edges in the computed DAG $\DAG$, resulting in the graph $\DAG_r$. 
    As visible from the pseudo-code in Alg.~\ref{alg:refine_alg}, we propose an iterative contestation process, that outputs a revised causal DAG and a NN adhering to it, when the SMEs agree with the computed DAG, given the constraints they imposed.
    
    \begin{table*}[h!t]
        \caption{Experiments with real data.
        We report mean MSE or AUC (std) for regression and classification, respectively, across different sample sizes of the training data ($N$) and 5-fold nested cross validation, best results in bold. Observed significance levels against CASTLE baseline are reported with the following intervals: 0 `***' 0.001 `**' 0.01 `*' 0.05 `.' 0.1 ` ' 1. NA indicates a data size bigger than the full dataset. We also detail the number of features/nodes $|V|$ and the number of edges $|E|$ in the injected DAG (for our method) and in the graph drawn from the extracted adjacency matrix (for CASTLE). \emph{Injected} columns refers to \S\ref{sec:real_full_dag}, \emph{Partial} to \S\ref{sec:partial_real} and \emph{Refined} to \S\ref{sec:refine_real}. 
        }
        \label{tab:real_castle_vs_inject}        
        \centering \fontsize{3mm}{3.5mm}\selectfont
        \begin{tabular}{rclll|cl|cl|cl}
        \toprule 
         & \multicolumn{6}{c|}{\textbf{CLASSIFICATION (Metric: AUC)}} & \multicolumn{4}{c}{\textbf{REGRESSION (Metric: MSE)}}\\ 
          & \multicolumn{4}{c|}{\textbf{Adult ($|V|=14$)}} & \multicolumn{2}{c|}{\textbf{HELOC ($|V|=23$)}} & \multicolumn{2}{c|}{\textbf{California ($|V|=8$)}} & \multicolumn{2}{c}{\textbf{Boston ($|V|=14$)}}  \\ 
        \!\! \textbf{Data}\!\! \!\! & \!\!\textbf{CASTLE}\!\! & \!\!\textbf{\textit{Injected}}\!\! & \!\!\textbf{\textit{Partial}}\!\!  & \!\!\textbf{\textit{Refined}}\!\! & \!\!\textbf{CASTLE}\!\! &\!\! \textbf{\textit{Injected}}\!\! & \!\!\textbf{CASTLE}\!\! & \!\!\textbf{\textit{Injected}}\!\! & \textbf{CASTLE}\!\! & \!\!\textbf{\textit{Injected}}\!\!\\
        \textbf{($N$)}& $|E|=210$ & $|E|=46$ & $|E|=116$ & $|E|=30$ & $|E|=552$ & $|E|=85$ & $|E|=72$ &$|E|=31$ & $|E|=182$ & $|E|=48$  \\
        \hline \vspace{-0.7em} \\ 
\!\!100   \!\!\!\!&  \!0.67 (0.03) \! &\!\textbf{0.69 .   }\! &\!        0.66    \!         &\!\textbf{0.69  . }\!  & \!\textbf{0.75 (0.02)}\! &  \!0.74       \! & \!\!7.05 (12.81)\!\!& \!\!\textbf{2.94  *** } \!\!& \!\!112.04 (91.06)\!\!& \!\textbf{86.17  *** }\!  \\
\!\!500   \!\!\!\!&  \!0.72 (0.04) \! &\!\textbf{0.74 *   }\! &\!        0.71    \!         &\!\textbf{0.74  * }\!  & \!\textbf{0.79 (0.01)}\! &  \!0.78   *** \! & \!\!2.33 (1.39)\!\! & \!\!\textbf{2.25    }   \!\!& \!\!21.95 (6.84)  \!\!& \!\textbf{20.45  * }\!   \\
\!\!1000  \!\!\!\!&  \!0.75 (0.03) \! &\!\textbf{0.76     }\! &\!        0.74    \!         &\!\textbf{0.76    }\!  & \!0.78 (0.01)\!          &  \!0.78       \! & \!\!2.96 (4.12)\!\! & \!\!\textbf{1.68  ** }  \!\!& \!\!NA\!\!  &  \!NA\!                               \\
\!\!2000  \!\!\!\!&  \!0.74 (0.03) \! &\!\textbf{0.77 *** }\!&\!\textbf{        0.76  *}\!&\!\textbf{0.77  *** }\!& \!\textbf{0.79 (0.01)}\! &  \!0.78   *** \! & \!\!3.86 (3.68)\!\! & \!\!\textbf{1.71  *** } \!\!& \!\!NA\!\!  &  \!NA\!                               \\
\!\!5000  \!\!\!\!&  \!0.75 (0.03) \! &\!\textbf{0.79 *** }\!&\!\textbf{        0.76   }\!&\!\textbf{0.79  *** }\!& \!0.79 (0.01)\!          &  \!0.79       \! & \!\!4.91 (7.41)\!\! & \!\!\textbf{1.51  *** } \!\!& \!\!NA\!\!  &  \!NA\!                               \\
\!\!10000 \!\!\!\!&  \!0.75 (0.02) \! &\!\textbf{0.85 *** }\!&\!\textbf{        0.76  .}\!&\!\textbf{0.85  *** }\!& \!\textbf{0.80 (0.01)}\! &  \!0.79   *** \! & \!\!1.74 (1.70)\!\! & \!\!\textbf{1.16  * }   \!\!& \!\!NA\!\!  &  \!NA\!                               \\
\!\!20000 \!\!\!\!&  \!0.76 (0.02) \! &\!\textbf{0.86 *** }\!&\!\textbf{        0.77  .}\!&\!\textbf{0.86  *** }\!& \!NA\!                   &  \!NA\!          & \!\!\textbf{0.66 (0.08)}\!\! &\!\!1.02  **  \!\!& \!\!NA\!\!  &  \!NA\!                               \\
        \hline
        \end{tabular}
    \end{table*}       
    SMEs are given the possibility to contest some or all of the relationships in the the DAG learnt from data and previous incorporated feedback, at any point of the learning process. This aims at assisting practitioners in validating and rectifying the causal relations discovered from the data, effectively debugging the NN based on its weights' structure.
    In the next section, we provide a case study on real data demonstrating how a practitioner can use our contesting algorithm to build more principled and predictive NNs.
    
\section{Empirical Evaluation}
\label{sec:experiments}
    We carried out two sets of experiments, on real and synthetic data, to assess the benefits of our proposed methods. 
    Firstly, we present a case study on real data (\S\ref{sec:realExp}), exemplifying the benefits that our methodology provides to modelling tasks in high-stakes decisions, when practitioners need to validate the relationships that the model is leveraging for its recommendations.    
    Additionally, we provide experiments with synthetic data (\S\ref{app:sythExp:gen}) which, in line with \cite{chowdhury_inducedeval,constantinou2021impact}, show that prior knowledge helps the causal discovery for low data regimes, further motivating the importance of causal knowledge injection. Details of the implementation, including code, are in Appendix~\ref{app:exp_details}% Supplementary Material (SM)
    .

\vspace{-0.3cm}
\subsection{Case Study with Real Data}
    \label{sec:realExp}
    We use real financial data from four publicly available datasets (see details in Appendix~\ref{app:data}% the SM
    ): two classification and two regression tasks. For the classifications, we use the Adult Income dataset~\cite{adultdata}, useful for affordability checks in the lending business to predict whether a loan applicant's income is greater than USD50K and the FICO HELOC data~\cite{FICODataset} for \emph{credit risk} assessment, to predict whether an applicant is likely to repay a loan.
    The two regression tasks are instead about predicting house prices: we use the Boston~\cite{bostondata} and California~\cite{calidata} Housing datasets. 
    We report analysis for three scenarios using Alg.~\ref{alg:refine_alg}: reconstructing a \emph{full} DAG via threshold $\tau$ optimisation, with no prior causal knowledge, to then inject it into our NNs (\S\ref{sec:real_full_dag}); injecting partial a priori causal knowledge expressing basic common sense assumptions for the Adult dataset (\S\ref{sec:partial_real}), and showcasing how practitioners can contest the DAG computed in \S\ref{sec:real_full_dag} using the assumptions adopted in \S\ref{sec:partial_real} (\S\ref{sec:refine_real}). In the absence of a true DAG for these datasets, our quantitative metric is predictive performance, namely MSE for regression and Area Under the Curve (AUC) for classification. All results are reported with observed significance levels for a two-tailed two-sample t-test (details of the testing procedure is reported in Appendix~\ref{app:detail_stat_tests}
    %the SM
    ). Our objective is demonstrating that predictive performance is not necessarily impacted by knowledge injection which can, in turn, help building more transparent and validated models.
 
\vspace{-0.3cm}
\subsubsection{\textbf{No a priori Knowledge}}
    \label{sec:real_full_dag}
    In this first scenario we assume no a priori causal knowledge, and use Alg.~\ref{alg:refine_alg} to construct potential DAGs by optimising the choice of threshold $\tau$, to then inject them into the NNs and measure the predictive performance with and without injection. Within the $\DAGmap_\tau$ function from Eq.~\ref{eq:direction_tau}, used at the beginning of Alg.~\ref{alg:refine_alg}, we tried between 10 and 15 different thresholds $\tau$ for each dataset and chose the ``best'' DAG through the evaluation of the change in predictive performance. For each dataset, we selected the DAG with lowest MSE/highest (AUC) and, as tie breaker, the lowest number of edges in the computed DAG.
    Details of this evaluation are in Appendix~\ref{app:real_tau_optim}%the SM
    .
    
    Within Alg.~\ref{alg:refine_alg}, we make the contesting process terminate with the DAGs chosen through this automatic procedure. 
    We aimed at checking whether fixing a plausible, though not humanly validated, DAG can lead to better predictive performance and whether the level of improvement depends on data size.
      
    As reported in Table~\ref{tab:real_castle_vs_inject}, 
    the predictive performance of the injected NNs can be up to 2.4x better than CASTLE (California with N=100). AUC for the Adult dataset is consistently above CASTLE (up to $13\%$ and significantly so for most of the sample sizes) while using only $\unsim20\%$ of the relationships that the ``unconstrained'' CASTLE network uses.\footnote{This reduction matches the reduction in NNs' weights at the input layer.} Similarly, for the Boston data, injection reduces edges by $\unsim75\%$ while significantly improving performance. For the California dataset, the MSE is significantly better for all sample sizes but the biggest and $N=500$ where there is not significant difference in the means. However, the reduction of computed DAG's edges is $\unsim40\%$. Finally, for the HELOC dataset the AUC for the injected NN is not significantly lower than CASTLE in half the cases, and by a maximum of 1\%, but with a much sparser NN: the amount of first layer's weights is only 15\% of the unconstrained NN. This stark reduction  will have included some useful relationships that, with some refinement, could be reintroduced to improve performance. We note that, by \emph{minimality}~\cite{pearl_2009} or \emph{parsimony}~\cite{vandekerckhove201514parsimony}, when the performance stays equal, a modeller should prefer the sparser, more parsimonious model. 
    Overall, simulating the use of Alg.~\ref{alg:refine_alg} without prior knowledge produced parsimonious NNs with an average $75\%$ less connections in the input layer. Moreover, predictive performance generally improves significantly or, in the worst case, worsens by at most 1\%, with no clear distinction by sample size.
  
    Note that, for this first scenario, our strategy for choosing the DAGs to inject is purely mechanic. 
    The adopted strategy is meant to gauge impact on predictive performance without a qualitative assessment of the validity of domain specific causal assumptions, a task that SMEs should carry out. 
    We envisage this strategy as a useful starting point in the absence of causal knowledge defined a priori. However, in real life applications, we intend the use of Alg.~\ref{alg:refine_alg} by a panel of SMEs, iteratively assessing intermediate outputs to refine the NN in light of previous experiments, leveraging their experience and knowledge of the modelling task, while learning more about it. 
    
    Next, we introduce two experiments providing  examples of how Alg.~\ref{alg:refine_alg} can work in a human-AI collaboration setting, with SMEs constraining and contesting DAGs computed by NN. We run these experiments only on the Adult dataset, as it contains some features, notably \emph{race, sex, age} and \emph{native-county}, that lend themselves to the construction of common sense causal assumption by lay users e.g., sex cannot be caused by age. This way we avoid formulating domain specific assumptions while providing a tangible example application.
    
    \begin{figure}[b] 
    \centering
        \includegraphics[width=0.45\textwidth, keepaspectratio]{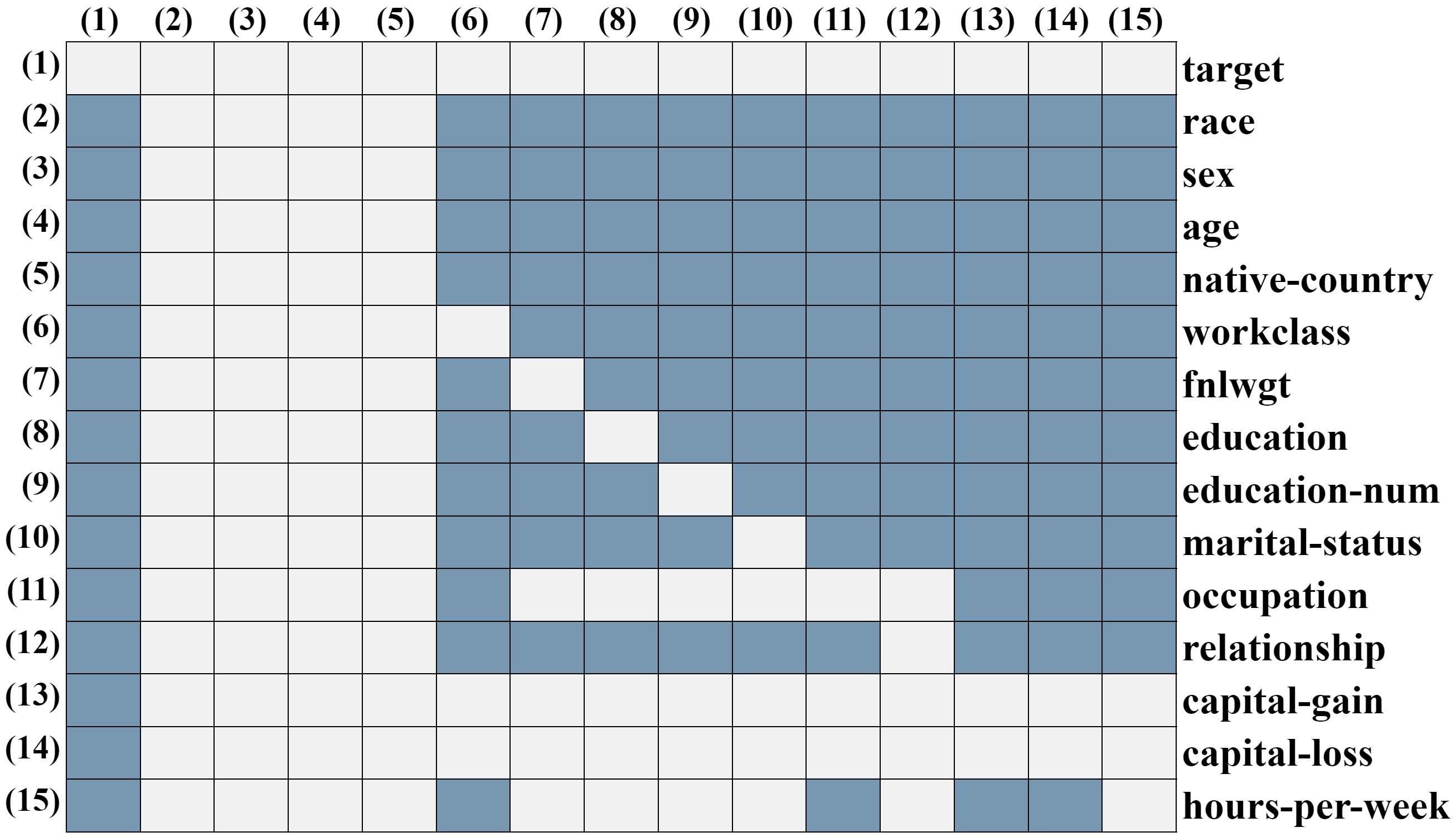}
        \caption{Input graph $\DAG_p$, as partial causal knowledge for the Adult dataset, in the form of an adjacency matrix \Adjmat{}. Blue represents edges; missing edges in white (hard constraints).}
        \label{fig:adult_ex-mat}
    \end{figure}           
    
    \vspace{-0.3cm}
\subsubsection{\textbf{Partial a priori Knowledge}}
    \label{sec:partial_real}
    In this experiment we build a very simple, yet intuitive, partial input graph $\DAG_p$ for the Adult dataset to constrain the NN to respect the following assumptions (reflected in the adjacency matrix in Fig.~\ref{fig:adult_ex-mat}% in the SM
    ):
    \begin{itemize}
        \item \emph{race, sex, age, native-county} cannot be caused by any feature, i.e. these features cannot have incoming edges. As a result, columns 2 to 5 of the adjacency matrix in Fig.~\ref{fig:adult_ex-mat} are blanked ($w_{ik}=0$);
        \item \emph{occupation} and \emph{hours-per-week} cannot cause \emph{fnlwgt (demographics index), education, education-num,  relationship} and \emph{marital-status}; the respective cells in Fig.~\ref{fig:adult_ex-mat} are blanked;
        \item the target ($income>USD50K$) cannot cause any feature and \emph{capital-gain, capital-loss} can only affect the target; rows 1, 13 and 14 in Fig.~\ref{fig:adult_ex-mat} are blanked.
    \end{itemize}
    Note that some of these assumptions could easily be confuted, e.g., by arguing that the target \emph{can} cause features, such as \emph{capital-gain} and \emph{loss}. We adopt these assumptions only to illustrate the effects on the adjacency matrix and to simulate a scenario whereby the modeller is testing whether the algorithm finds relationships in the data that help the prediction task. As in \cite{pearl_2009}, we believe that the opportunity to extract and enforce such assumptions, or simply talk about them, has the potential to make models more transparent, robust and representative of causal mechanisms of the world.
      
    We report the AUC of the NN injected with the graph in Fig.~\ref{fig:adult_ex-mat} in Table~\ref{tab:real_castle_vs_inject}, \emph{\textbf{Partial}} column.
    The ``causal constraints'' 
    result in performances generally not significantly different from CASTLE, but with a computed DAG that has about a half the edges of the unconstrained NN ($|E|=116$ vs $|E|=210$). On the whole, we obtain a sparser NN, adherent to common-sensical causal knowledge following our assumptions, whose recommendations are therefore arguably more understandable and trustworthy, and whose performance is comparable to an unconstrained NN.

\vspace{-0.3cm}
\subsubsection{\textbf{Contesting a Computed DAG}}
    \label{sec:refine_real} 
\begin{figure}[ht] 
        \centering
        \includegraphics[width=0.43\textwidth, keepaspectratio]{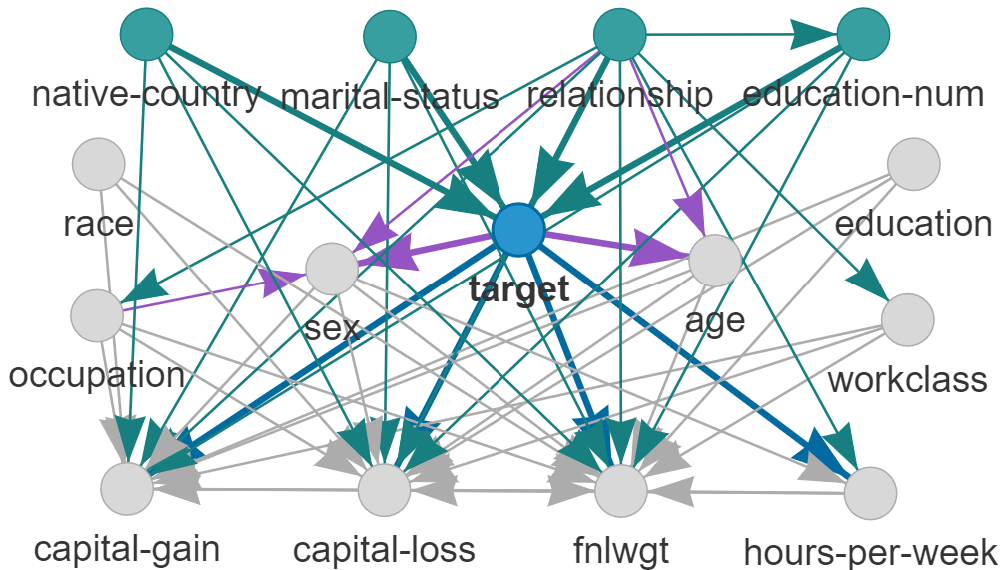}
        \caption{Example of computed DAG for Adult dataset. 
        Cyan nodes at the top are computed causes for the target (``Income>50K''), edges coming out of the target are in blue while in purple are the edges into nodes that cannot be caused (as per basic assumptions in Fig.~\ref{fig:adult_ex-mat}).
        } 
        \label{fig:adult_ex-dag}
    \end{figure}

    Our last experiment on real data aims at showcasing the process of contesting the causal structure computed and used by the NN, as afforded by Alg.~\ref{alg:refine_alg}. The experiment starts, as the experiment in \S\ref{sec:real_full_dag}, with no prior knowledge about the problem. We adopt the same threshold optimisation strategy detailed in Appendix~\ref{app:real_tau_optim} 
    %the SM 
    and chose $\tau=0.08$. Hence, in the first few runs of Alg.~\ref{alg:refine_alg} the output of the \texttt{revise\textunderscore graph} function would only change the threshold $\tau$. 
    Having chosen $\tau=0.08$, the DAG is extracted and shown to the SMEs, as in
    Fig.~\ref{fig:adult_ex-dag}. 
    In this DAG, the target (Income>50K) is deemed to be causing a few features including sex and age (see purple arrows).     
       
     \begin{figure*}[h!t]
        \centering
        \includegraphics[width=\textwidth]{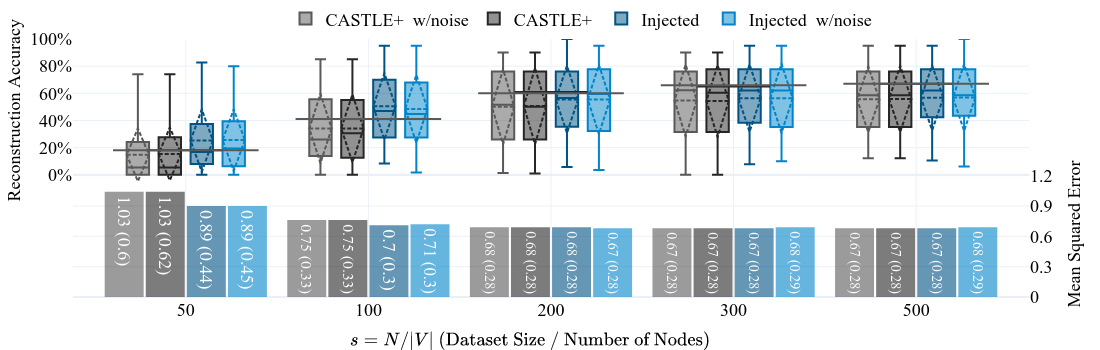}
	    \caption{ Reconstruction Accuracy and MSE when changing $s=N/|V| \in \{50,100,200,300,500\}$, the sample size $N$ proportional to the number of nodes $|V| \in \{10, 20, 50\}$ in the causal DAG \DAG. 
	Darker (grey or blue) colors refer to the no-noise scenario, whereas lighter colors refer to the scenario with noise.
	The values are an average over 10 runs for each combination of $|V|, s \text{ and } e=|E|/|V| \in \{1,2,5\}$. 
        The boxplots (left y axis) show Min/Max/Median (solid lines) and Mean/Std (dashed lines) of the reconstruction accuracy. The bottom bars (right y axis) show the MSE (std). 
        The solid horizontal lines spanning across each of the pairs  of boxplots are the re-based value of the CASTLE+ mean to account for the advantage that our Injection methodology knows 20\% of the edges. If the mean of \emph{Injected} is above/below the level of the horizontal lines, the average increase in reconstruction accuracy is more/less than proportional to the amount injected. }
        \label{fig:alphaplot_noise}
    \end{figure*}       
    This is counter-intuitive from a common sense, let alone causal, perspective.
    Thus, the contestation now addresses specific edges in the computed DAG, and \texttt{revise\textunderscore graph} produces a DAG $\DAG_r$ that differs from  \DAG{} in Fig.~\ref{fig:adult_ex-dag} by the purple edges. $\DAG_r$ is then injected back into the NN producing the results in Table~\ref{tab:real_castle_vs_inject}, \emph{\textbf{Refined}} column: the NN injected with the DAG refined by means of contesting is not only more intuitive and adhering to common sense, but presents the same predictive performance as the NN using non-sensical relationships and significantly better predictive performance than CASTLE. 
    Ultimately, our \emph{Refined} NN is 7x smaller in the input layer, adheres to common sense, and yet it is up to 13\% more predictive than an unconstrained NN.

    \vspace{-0.3cm}
    \subsection{Experiments with Synthetic Data }
    \label{app:sythExp:gen}

    To confirm the results from our case study on real data, we investigate the effectiveness of our proposed method on synthetic data. 
    Our simulations compare scenario (i) in \S\ref{sec:real_full_dag}, where no prior causal knowledge is available, to scenario (ii) in \S\ref{sec:partial_real}, where practitioners do have a set of a priori assumptions. The comparison of scenarios (i) and (ii) can be seen, in the setting with synthetic data serving as a proxy for domain experts, as amounting to the scenario in \S\ref{sec:refine_real}, where contesting an initial computed DAG corresponds to providing a priori knowledge in scenario (ii); the only difference lies in the starting point. We chose to test these scenarios also because of the easier simulation.
    The experiments aim at answering the following questions:  
(Q1) Does knowledge injection by our algorithms improve predictive performance?  
(Q2) Can we reconstruct a DAG, known to be underpinning the DGP, using Alg.~\ref{alg:refine_alg}? 
(Q3) How well can Alg.~\ref{alg:refine_alg} fill the gaps of an input graph contributing only partial knowledge? 
(Q4) How does knowledge injection performance change in different data size regimes?
(Q5) How resilient are our algorithms to noise?
   
    Using Alg.~\ref{alg:refine_alg}, we represent the \emph{Expert Knowledge} with an input graph $\DAG$ encapsulating a priori partial causal knowledge among a subset of the features fed to the NN.
    In the experiments shown in Fig.~\ref{fig:alphaplot_noise} we inject a 20\% 
    random sample of the total amount of edges in the true DAG; experiments with 10\% and 50\% of DAG edges injected are reported in Appendix~\ref{app:known edges}%in the SM
    .  Once the known edges are selected, the entries of the edges representing the opposite direction in the adjacency matrix \Adjmat{} are set to 0, e.g. $X_i\rightarrow X_j$ is selected as known, then $(i,j) \in E$ and $w_{ji}=0$.     
    
\textit{\textbf{Synthetic Data Generation}.}  
    We generate synthetic data adhering to a series of randomly generated DAGs of different sizes, using the methodology of \cite{CASTLE}.\footnote{For details  refer to Appendix B.1 of \cite{CASTLE}. Note that we standardise all feature values in the generated data to  mean 0 and std 1. Thus, following \cite{reisach2021beware}, our results can be regarded as conservative estimates of reconstruction accuracy. }
    The generated synthetic DAGs and data vary across three main dimensions: number  of nodes in \DAG{} ($|V| \in \{10, 20, 50\}$), number of edges ($|E| = |V|*e$, where $e \in \{1,2,5\}$), and data size ($N=|V|*s$, where $s \in \{50,100,200,300,500\}$). In the remainder, we refer to $s$ as \emph{proportional sample size}. 
    
\textit{\textbf{Evaluation Metrics}.}  
    \label{app:baseline}
    We use average and Standard Deviation (Std) of \emph{Mean Squared Error (MSE)} for the evaluation of \emph{predictive performance} for questions Q1, Q4 and Q5. For the evaluation of \emph{reconstruction accuracy} for questions Q2, Q3 and Q5 we use the distribution of the percentage of edges that match those in the true DAG. We run each scenario, involving one of the combinations of $|V|,\text{ }e \text{ and } s$, 10 times and report the distribution of results as a boxplot in Fig.~\ref{fig:alphaplot_noise}, in comparison with the baselines detailed next. As in the experiments with real data, differences in means are tested with a t-test.
    
\textit{\textbf{Baselines}.}  
    CASTLE's predictive performance has been compared to the main NNs' regularisation methods in the literature \cite{CASTLE}, providing the best performance for both classification and regression tasks on both synthetic and real data. Hence, for prediction, our baseline is a well-performing regularised NN: CASTLE. 
    However, CASTLE alone cannot be used as a baseline for reconstruction accuracy but the adjacency matrix extracted from CASTLE using Eq.~\ref{eq:adjmat} provides a useful starting point. Our baseline is thus built by: using CASTLE's weights \weightset{} to compute $\Adjmat_\weightset$ according to Eq.~\ref{eq:adjmat} and then Eq.~\ref{eq:direction_tau} to derive a DAG $\DAG = \DAGmap_\tau(\Adjmat_\weightset)$. We call this method CASTLE+.
    
    The differences between CASTLE+ and our Alg.~\ref{alg:inject_alg} are as follows. For CASTLE+ we let CASTLE run unconstrained and only after training we extract an adjacency matrix (Eq.~\ref{eq:adjmat}) and transform it into a DAG (Eq.~\ref{eq:direction_tau} with an appropriate choice of threshold $\tau$). Instead, our method for injecting the DAG involves feeding a graph into the NN so that a mask is applied and only the non-masked weights are optimised. For an illustration of the procedure as to how we inject causal knowledge through masking, refer to \S\ref{sec:castle}. Note that CASTLE+ amounts to using Alg.~\ref{alg:inject_alg} with a complete graph $\DAG_p$.
    In the experiments we use the value of $\tau$ that produced the lowest number of mismatches, for each of CASTLE+ and our method.\\

    \textit{\textbf{Simple DGP}}. The results for this scenario are given in darker colors in Fig.~\ref{fig:alphaplot_noise}, where we show two metrics: the predictive performance (MSE) with the bars at the bottom of the figure; and the distributions of reconstruction accuracy through the boxplots at the top of the figure. The results presented vary across one of the three dimensions mentioned earlier, namely the proportional sample size $s$. Analysis of the changes over the other two dimensions ($e = |E|/|V|$ and $|V|$) are left to 
    Appendices~\ref{app:edges} and~\ref{app:nodes}%the SM
    .
    From Fig.~\ref{fig:alphaplot_noise} we can observe that the predictive performance improves with injection for small data regimes (up to 15\% for $s<200$) while it is not affected for bigger proportional sample sizes. However, none of the effects are statistically significant. Also for reconstruction accuracy the biggest gains are again observed in the low data scenario. The proportional gains in the number of correct edges is greater than the proportion of edges injected by up to 10\%, as represented by the mean of the boxplots lying above the gray longitudinal lines spanning across them. The increases for $s=50$ and $s=100$ result to be statistically significant at the 5\% level ($t(178)=2.226, p=0.027$ and $t(178)=2.464, p=0.015$, respectively). For the other $s$, no significant differences are observed.
    With these experiments we can answer questions Q1 through Q4: Alg.~\ref{alg:inject_alg} can improve DAG reconstruction (Q2) as well as fill in gaps in partial causal knowledge (Q3), with no decrease in prediction performance (Q1), but only for low data regimes (Q4).
        
    \textit{\textbf{Noisy DGP}}. To assess the robustness of Alg.~\ref{alg:inject_alg} to noise, aiming at answering question Q5, we add to the training data a number of features amounting to 20\% of the number of nodes in the DAG used to generate the data. These additional ``noisy'' features are generated out of a standard normal distribution and have no links to the other features in the data. Results are again presented in Fig.~\ref{fig:alphaplot_noise} for ease of comparison with the no-noise scenario. As visible from the bottom bar charts, the MSE for the target feature $Y$ stays effectively the same across the different proportional data sizes (no significant differences in mean). Also the reconstruction accuracy (top boxplots) appears not to be affected at all (again, no significant differences in mean). This is in line with the results presented in \cite{CASTLE}. We can conclude that our algorithm is resilient to noise with regard to both reconstruction and predictive performance (Q5).

    \vspace{-0.4cm}
    \section{Conclusion}
        \label{sec:conclusion}
    \vspace{-0.2cm}
    The proposed methods represent a principled approach to fitting neural networks (NNs): we leverage knowledge injection in the form of causal graphs to empower technical experts to contest NNs, based on the structural assumptions discovered from the data.
    We propose two algorithms to deliver \emph{contestable}
    NNs: the first unlocks contestability by allowing networks to take feedback in the form of causal graph injection; the second uses computed causal graphs to elicit feedback from experts, so that they can contest the data-driven causal graph and inject their causal views into the NN. 
    We apply our algorithms to real financial datasets demonstrating how they can yield very parsimonious, hence more interpretable and easier to debug, NNs, while either significantly improving or losing very little predictive performance.
    Finally, we demonstrate, through empirical results on synthetic data, that knowledge injection, as afforded by our method, generally improves causal discovery in low data regimes, despite noise. 
    We used predictive performance to assess the viability of our method for prediction tasks and found that knowledge injection can produce similar or better performing NNs, but with the added confidence of being able to explore the relationships used in the predictions.
    
    We envisage interesting lines of future work including:
    exploring the indirect causal effects that take place in the hidden layers of our injected NN;    introducing Bayesian learning weight updates, e.g. as in \cite{mullachery2018bayesian}, to improve our method's human-AI collaboration capabilities through uncertainty quantification. 
    Further, we would like to equip our method with the capability to aggregate independent views of several experts into a consistent causal graph, similar to Alrajeh et al.~\cite{alrajeh2020combining}'s proposal for causal models, and to allow the enforcement of the presence of causal direction on top of the absence thereof.
    Our proposed method not only produces contestable NNs, but also improves their interpretability and, we believe, their trustworthiness. This is because SMEs are called to examine computed causal graph and provide feedback that the NN is guaranteed to respect: we will explore this angle in future work.
    Finally, we plan to explore further the human-in-the-loop debugging capabilities of our method, especially for high-stakes decision models, and conduct user studies on the propensity and efficacy of SMEs in understanding and contesting model outputs depending on presentation and interaction modalities.

\ack We would like to thank Ruben Menke, Torgunn Rings{\o}, Antonio Rago, Francesco Leofante, Mark Somers and all the anonymous reviewers for the helpful feedback on earlier version of the paper. Russo was supported by UK Research and Innovation (grant number EP/S023356/1), in the UKRI Centre for Doctoral Training in Safe and Trusted Artificial Intelligence (\url{www.safeandtrustedai.org}). Toni was partially funded by the ERC under the EU’s Horizon 2020 research and innovation programme (grant agreement No. 101020934) and by J.P. Morgan and by the Royal Academy of Engineering under the Research Chairs and Senior Research Fellowships scheme. 

\bibliography{main}

\appendix
\newpage
\noindent
\textbf{\Large{Appendices}}
\setcounter{section}{0}

\renewcommand{\thesection}{\Alph{section}}

\section{Further Details for Experiments }
\label{app:exp_details}
 
    For all  experiments\footnote{Our code is available at:  \url{https://github.com/briziorusso/causal-injection-FFNN/tree/main}} we choose 3 hidden layers ($M=3$) of sizes $2*|V|, 2/3*|V|, 2*|V|$ respectively, with ReLU activations. Experiments with smaller networks are provided for synthetic data in \S\ref{app:netsize}. All networks are initialized and seeded identically and use the Adam optimizer with a learning rate of 0.001 for a maximum of 1000 steps ($T$). A patience ($T_s$) of 50 steps on the loss on the validation set is used to stop training. 
    Results for the synthetic dataset are then reported for 
    10 randomly generated DAGs. For the real data experiments, given the hyper-parameters optimisation of the threshold $\tau$, we have used 5-fold nested cross validation and the results reported are the average over the resulting 25 runs.

\subsection{Experiments with Real Data}
\label{app:data}
Here we provide additional details for the experiments in \S\ref{sec:realExp}: we start from the datasets to then cover the optimisation of the threshold $\tau$, part of experiment (i) in \S\ref{sec:real_full_dag}.

We report details for the datasets in Table~\ref{tab:real_data_det}. For the regression tasks, with Boston~\cite{bostondata} and California Housing~\cite{calidata}, we used the data out of the box from scikit-learn\footnote{\url{https://scikit-learn.org/stable/datasets/toy_dataset.html}}. For the classification tasks, with HELOC\footnote{Data and pre-processing mapping (keys.csv) is downloadable from our repo at \url{https://github.com/briziorusso/causal-injection-FFNN/tree/main/data/fico/};}~\cite{FICODataset} and 
Adult Income\footnote{Data and pre-processing mapping are from Penn Machine Learning Datasets repository at \url{https://github.com/EpistasisLab/pmlb/blob/master/datasets/adult/}}~\cite{adultdata}, we used pre-processed data. Additionally, for Adult, we sampled 20000 observations and we balanced the proportion of positive and negative examples in target feature to 50-50 (from 25-75).  
% \vspace{-1cm}
\begin{table}
        \caption{Real-world dataset details. Type is Regression (R) for real-valued target or Classification (C) for binary target.}
        \label{tab:real_data_det}
        \centering 
        \begin{tabular}{ >{\arraybackslash}m{3.3cm} cccc}
       \textbf{Dataset}  & \textbf{Sample Size}  & \textbf{Features} &\textbf{Type}\\
        \hline\\[-0.8em]
Boston Housing (BH)                    & 506   & 14 & R\\
California Housing (CH)                & 16512 & 8  & R\\
Home Equity Line Of Credit (HELOC)     & 7844  & 23 & C\\
Adult Income (IN)                      & 48842 & 14 & C\\
            \hline
        \end{tabular}
\end{table} 

\subsubsection{Details on Statistical Tests}
\label{app:detail_stat_tests}
Here we report the details to reproduce the observed significance levels reported in Table~\ref{tab:real_castle_vs_inject}. Table~\ref{tab:real_castle_vs_inject_std} reports the means and standard deviations for all the experiments. As mentioned in the main text, we used a two-tailed, two-sample t-test for difference in means. Each sample was made up of 25 runs within a 5-fold nested cross validation resulting in 48 degrees of freedom for the Student's t-distribution. The t-statistics and associated p-values are reported in Table~\ref{tab:real_castle_vs_inject_tests}. Note that for California and Boston the homoscedasticity (equal variance) assumption was not satisfied (after testing the difference in variance with an F-test) hence we used the heteroscedastic t-test.

\begin{figure}[h!t]   
    \subfloat[Adult Income]{
    \includegraphics[width=0.97\linewidth, keepaspectratio]{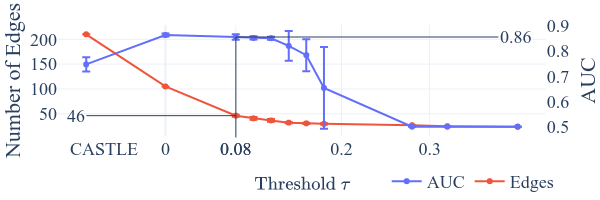}
    \label{fig:adult_tau}}\\[-0.3em]     
    \subfloat[FICO HELOC]{
    \includegraphics[width=0.97\linewidth, keepaspectratio]{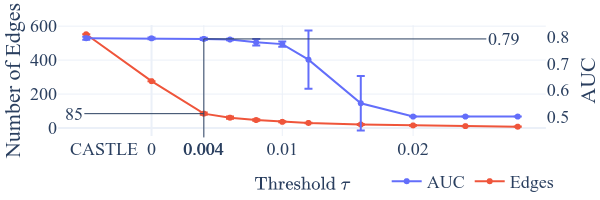}
    \label{fig:fico_tau}}\\[-0.3em]  
    \subfloat[Boston Housing]{
    \includegraphics[width=0.97\linewidth, keepaspectratio]{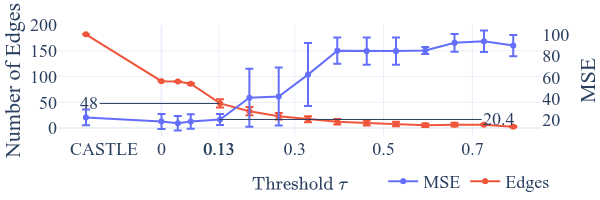}
    \label{fig:boston_tau}}\\[-0.3em]  
    \subfloat[California Housing]{
    \includegraphics[width=0.97\linewidth, keepaspectratio]{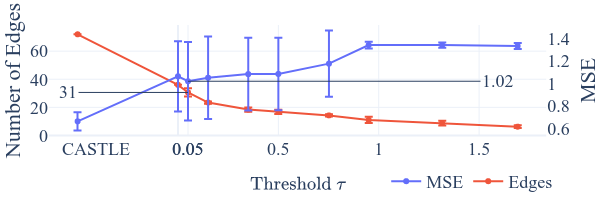}
    \label{fig:cali_tau}}\\[-0.3em]  
    \caption{
    Threshold $\tau$ optimisation for experiment in \S\ref{sec:real_full_dag}. Here $\tau <0$ corresponds to the application of ``unconstrained'' CASTLE \cite{CASTLE}. Chosen thresholds are in bold on the x-axis. The number of edges and the predictive performance of the network injected with the DAG derived with the chosen $\tau$ are reported on the y-axes.\\[-0.5cm] }
    \label{fig:tau_optims}
\end{figure}     

\subsubsection{Threshold Optimisation.}
\label{app:real_tau_optim}
    Here we provide details on the optimisation of the threshold $\tau$ to chose a DAG without having to make qualitative causal judgments, as part of the experiment in \S\ref{sec:real_full_dag}. The results for all datasets are in Fig.~\ref{fig:tau_optims}. The optimisation runs through several thresholds and compares the number of edges (red) and the appropriate predictive performance metric 
    (blue) when increasing the threshold $\tau$, along the $x$ axis, so that more and more edges are masked. As expected, the number of edges decreases monotonically for bigger thresholds while MSE/AUCs are the trends of interest. 
    
    As visible in Fig.~\ref{fig:adult_tau} for the Adult dataset, $\tau=0.08$ keeps the best AUC, while reducing the number of edges by more than 50\% compared to $\tau=0$. Increasing $\tau$ produces small gains in parsimony (lower $|E|$) but starts to reduce performance. The same applies to the FICO HELOC data in Fig.\ref{fig:fico_tau}, where we select $\tau=0.004$, before the AUC starts to decrease. 
    For both datasets, valid alternative choices are $\tau=0.1$ and $0.01$, respectively, but we preferred a lesser worsening of performance for small gains in parsimony. 
    For the Boston dataset in Fig.\ref{fig:boston_tau}, we select $\tau=0.13$ as
    the MSE thereafter increases significantly. Finally, California (Fig.\ref{fig:cali_tau}) shows a different scenario: injecting always hurts predictive performance. We chose $\tau=0.05$ which has the lowest MSE. However, as visible in Table~\ref{tab:real_castle_vs_inject} in the main text, the worse MSE is not observed for injected NNs on smaller sample sizes.

       \begin{table*}[h!t]
        \caption{MSE or AUC (std) for regression and classification, respectively, across different sample sizes of the training data ($N$).
        }
        \label{tab:real_castle_vs_inject_std}        
        \centering \fontsize{2.8mm}{3.5mm}\selectfont
        \begin{tabular}{rcccc|cc|cc|cc}
        \toprule 
         & \multicolumn{6}{c|}{\textbf{CLASSIFICATION (Metric: AUC)}} & \multicolumn{4}{c}{\textbf{REGRESSION (Metric: MSE)}}\\ 
          & \multicolumn{4}{c|}{\textbf{Adult}} & \multicolumn{2}{c|}{\textbf{HELOC}} & \multicolumn{2}{c|}{\textbf{California}} & \multicolumn{2}{c}{\textbf{Boston}}  \\ 
        \!\! \textbf{Data}\!\! \!\! & \!\!\textbf{CASTLE}\!\! & \!\!\textbf{\textit{Injected}}\!\! & \!\!\textbf{\textit{Partial}}\!\!  & \!\!\textbf{\textit{Refined}}\!\! & \!\!\textbf{CASTLE}\!\! &\!\! \textbf{\textit{Injected}}\!\! & \!\!\textbf{CASTLE}\!\! & \!\!\textbf{\textit{Injected}}\!\! & \textbf{CASTLE}\!\! & \!\!\textbf{\textit{Injected}}\!\!\\
        \hline \vspace{-0.7em} \\ 
\!\!100   \!\!\!\!&  \!0.67 (0.03) \! &\!0.69 (0.04) \! &\!0.66 (0.02)\!&\! 0.69 (0.04) \!& \!0.75 (0.02)\! &  \!0.74 (0.04)\! & \!\!7.05 (12.81)\!\!         & \!\!2.94 (2.63)\! \!& \!\!112.04 (91.06)\!\!& \!86.17 (13.75)\!    \\
\!\!500   \!\!\!\!&  \!0.72 (0.04) \! &\!0.74 (0.02) \! &\!0.71 (0.02)\!& \!0.74 (0.02) \!& \!0.79 (0.01)\! &  \!0.78 (0.01)\! & \!\!2.33 (1.39)\!\!          & \!\!2.25 (1.07)\! \!& \!\!21.95 (6.84)  \!\!& \!20.45 (5.12)\!     \\
\!\!1000  \!\!\!\!&  \!0.75 (0.03) \! &\!0.76 (0.03) \!&\!0.74 (0.03)\! & \!0.76 (0.02) \!& \!0.78 (0.01)\!          &  \!0.78 (0.01)\! & \!\!2.96 (4.12)\!\! & \!\!1.68 (1.14)\! \!& \!\!NA\!\!  &  \!NA\!                        \\
\!\!2000  \!\!\!\!&  \!0.74 (0.03) \! &\!0.77 (0.01)\!&\! 0.76 (0.03) \!&\!0.77 (0.02) \! & \!0.79 (0.01)\! &  \!0.78 (0.01)\! & \!\!3.86 (3.68)\!\!          & \!\!1.71 (0.57)\! \!& \!\!NA\!\!  &  \!NA\!                        \\
\!\!5000  \!\!\!\!&  \!0.75 (0.03) \! &\!0.79 (0.03)\!&\! 0.76 (0.02) \!&\!0.79 (0.03) \! & \!0.79 (0.01)\!          &  \!0.79 (0.01)\! & \!\!4.91 (7.41)\!\! & \!\!1.51 (0.62)\! \!& \!\!NA\!\!  &  \!NA\!                        \\
\!\!10000 \!\!\!\!&  \!0.75 (0.02) \! &\!0.85 (0.01)\!&\! 0.76 (0.02) \!&\!0.85 (0.01) \! & \!0.80 (0.01)\! &  \!0.79 (0.01)\! & \!\!1.74 (1.70)\!\!          & \!\!1.16 (0.31)\! \!& \!\!NA\!\!  &  \!NA\!                        \\
\!\!20000 \!\!\!\!&  \!0.76 (0.02) \! &\!0.86 (0.01)\!&\! 0.77 (0.02) \!&\!0.86 (0.01) \! & \!NA\!                   &  \!NA\!                                & \!\!0.66 (0.08)\!\! &  \!\!1.02 (0.35)\!\!& \!\!NA\!\!  &  \!NA\!  \\
        \hline 
        \end{tabular}
    \end{table*}    

    \begin{table*}[h!t]
        \caption{t-statistic (p-value) comparing each column of Table~\ref{tab:real_castle_vs_inject_std} against the respective CASTLE baseline for each dataset and sample size.
        }
        \label{tab:real_castle_vs_inject_tests}        
        \centering \fontsize{2.8mm}{3.5mm}\selectfont
        \begin{tabular}{rccc|c|c|c}
        \toprule 
          & \multicolumn{3}{c|}{\textbf{Adult}} & \multicolumn{1}{c|}{\textbf{HELOC}} & \multicolumn{1}{c|}{\textbf{California}} & \multicolumn{1}{c}{\textbf{Boston}}  \\ 
        \!\! \textbf{Data}\!\! \!\! \!\! & \!\!\textbf{\textit{Injected}}\!\! & \!\!\textbf{\textit{Partial}}\!\!  & \!\!\textbf{\textit{Refined}}\!\!  &\!\! \textbf{\textit{Injected}}\!\!  & \!\!\textbf{\textit{Injected}}\!\!  & \!\!\textbf{\textit{Injected}}\!\!\\
        \hline \vspace{-0.7em} \\ 
\!\!100   \!\!\!\!&  \!2.000 (0.051) \!  &\! 1.387 (0.172) \! &\! 2.000  (0.051)  \!&\! 1.118 (0.269)  \! & \!1.571 (0.123) \! &  \!1.405 (0.167) \\
\!\!500   \!\!\!\!&  \!2.236 (0.030) \!  &\! 1.118 (0.269) \! &\! 2.236  (0.03)   \!& \!3.536 (0.001)  \! & \!0.228 (0.821) \! &  \!0.878 (0.384) \\
\!\!1000  \!\!\!\!&  \!1.179 (0.224) \!  &\! 1.179 (0.244) \!&\!  1.387  (0.172)  \!& \!0.000 (1.000)  \! & \!1.497 (0.141) \! &  \!NA            \\
\!\!2000  \!\!\!\!&  \!4.743 (0.000) \!  &\! 2.357 (0.023) \!&\!  4.160  (0.000)   \!&\!3.536 (0.001)  \! & \!2.887 (0.006) \! &  \!NA           \\
\!\!5000  \!\!\!\!&  \!4.714 (0.000) \!  &\! 1.387 (0.172) \!&\!  4.714  (0.000)   \!&\!0.000 (1.000)  \! & \!2.286 (0.027) \! &  \!NA            \\
\!\!10000 \!\!\!\!&  \!22.361 (0.000) \! &\! 1.768 (0.083) \!&\!  22.361 (0.000)   \!&\!3.536 (0.001)  \! & \!1.678 (0.100)   \! &  \!NA            \\
\!\!20000 \!\!\!\!&  \!22.361 (0.000) \! &\! 1.768 (0.083) \!&\!  22.361 (0.000)   \!&\!NA             \! & \!5.014 (0.000)     \! &  \!NA           \\
        \hline
        \end{tabular}
    \end{table*}

\subsection{Parameter Study for Synthetic Data}
    In \S\ref{app:sythExp:gen} we report results for a fixed number of edges injected (20\%). Here we provide comparisons of performance for different percentages of edges injected (\S\ref{app:known edges}).
    Moreover, the results reported in Fig.~\ref{fig:alphaplot_noise} show variation wrt one of the three dimensions considered when generating the random DAGs and data, namely, proportional sample size ($s=N/|V|$). Here, we report additional results for the other two dimensions: number of nodes $|V|$ (\S\ref{app:nodes}), 
    and proportion of edges over nodes $e=|E|/|V|$ (\S\ref{app:edges}). 
    Finally, we report a comparison for network size (\S\ref{app:netsize}). 
    The results of this section corroborate the ones presented in \S\ref{app:sythExp:gen}.

\subsubsection{Percentage of Known Edges.}
\label{app:known edges}
\begin{figure}[h!t]
    \centering
    \includegraphics[width=0.48\textwidth]{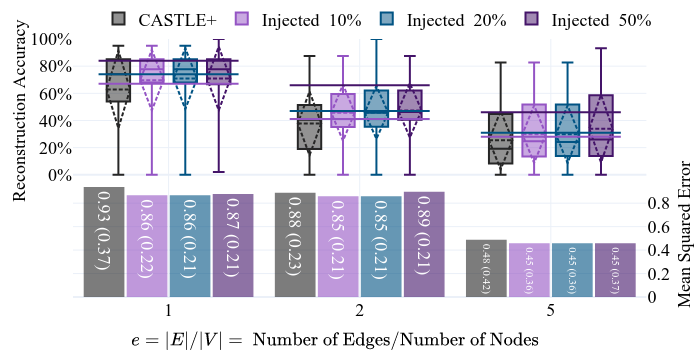}
    \caption{Reconstruction Accuracy and MSE when changing the proportion of edges over nodes (see \S\ref{app:edges}). The amount of known edges injected  is 10\%, 20\% (as in the paper, see Fig.\ref{fig:alphaplot_noise}) and 50\% (see \S\ref{app:known edges}).}
    \label{fig:edges_50percentknown}

\end{figure}

In Fig.~\ref{fig:edges_50percentknown} we vary the proportion of edges injected (10\%, 20\% as in Fig.\ref{fig:alphaplot_noise}, and 50\%). 
As visible, injecting 50\% of edges never pays off proportionally, i.e. the average reconstruction accuracy, although higher than CASTLE+, generates less of an increase than ``rebased'' CASTLE+ (solid horizontal lines). On the other hand, the reconstruction accuracy when injecting 10\% of the DAGs is always more than proportional to the injected amount. Most of the gains are recorded for denser DAGs ($e \in \{2,5\}$) when reconstruction is generally more difficult, as shown by the decreasing overall trend.
\subsubsection{Number of Edges in the DAG}
\label{app:edges}
    The effect of changing the proportion of edges per node ($e = |E|/|V|$) is presented in Fig.~\ref{fig:edges_50percentknown} (where CASTLE+ vs Injected 20\% is the scenario shown in Fig.\ref{fig:alphaplot_noise} in the main text).
    Results show that the sparser the DAG (the smaller $e$) the better the performance of our algorithm, 
    achieving reconstruction accuracy averaging at around 75\% for $e=1$. For $e=2$ the average drops to the average level across all proportional sample sizes $s$, 
    while increasing $e$ further, to 5, results in averages dropping to less then 40\%, comparable to the effect of having only 50 observations per node ($s=50$, Fig.\ref{fig:alphaplot_noise}). Overall, the denser the DAG, the worse the performance of both CASTLE+ and our algorithms.
\begin{figure}[h!t]
    \centering
    \includegraphics[width=0.48\textwidth]{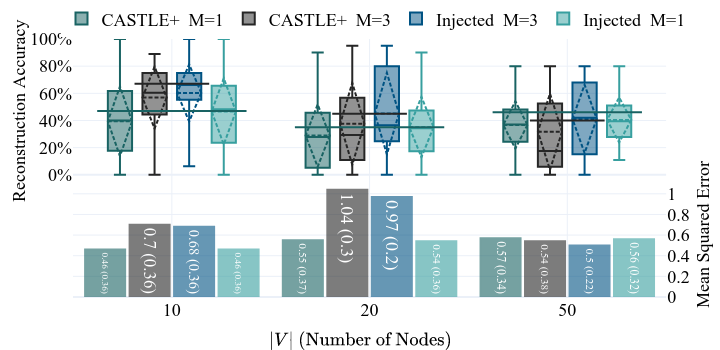} 
    \caption{Reconstruction Accuracy and MSE when changing the number of nodes in the DAG underpinning the data (see \S\ref{app:nodes})
    and the numbers of layers (M=3 % as in the paper, see 
    in Fig.\ref{fig:alphaplot_noise}, see \S\ref{app:netsize}).
    }
    \label{fig:netsize}

\end{figure}
\subsubsection{Number of Nodes in the DAG}
\label{app:nodes}
    MSE and reconstruction accuracy results when changing $|V|$ are shown in Figure \ref{fig:netsize} ($M=3$ correspond to the main scenario, as in Fig.\ref{fig:alphaplot_noise}). We can observe that the performance varies significantly for increasing DAG sizes. For $|V|=10$, both CASTLE+ and our method show better reconstruction accuracy than the average across $s$ (in Fig.~\ref{fig:alphaplot_noise}). For $|V|>10$, however, we observe a significant drop in overall performance for CASTLE+, whereas our method suffers less from the increased size of the DAG.
\subsubsection{Network Size}
\label{app:netsize}
    In Fig.~\ref{fig:netsize}, jointly with the analysis on the number of nodes, we show a comparison of 3-layers networks (used for the experiments in \S\ref{app:sythExp:gen}) with smaller networks of one single hidden layer and an amount of neurons of 3.2 times the number of input features (i.e. $M=1, h=(d+1)*3.2$).
    Interestingly, as visible from the bottom bar charts, the MSE for the target variable $Y$ generally improves with smaller networks, while the same change worsens reconstruction accuracy. Better prediction does not always couple with better causal discovery.

\end{document}